\documentclass[10pt, reqno, letterpaper, twoside]{amsart}
\usepackage[margin=1in]{geometry}

\usepackage{amssymb, bm, mathtools}
\usepackage[usenames,dvipsnames,svgnames,table]{xcolor}
\usepackage[pdftex, xetex]{graphicx}
\usepackage{enumerate, setspace}
\usepackage{float, colortbl, tabularx, longtable, multirow, subcaption, environ, wrapfig, textcomp, booktabs}
\usepackage{pgf, tikz, framed, url, hyperref}
\usepackage[normalem]{ulem}
\usetikzlibrary{arrows,positioning,automata,shadows,fit,shapes}
\usepackage[english]{babel}
\hypersetup{
    colorlinks,
    linkcolor={red!50!black},
    citecolor={blue!50!black},
    urlcolor={blue!80!black}
}\usepackage{microtype}
\microtypecontext{spacing=nonfrench}

\usepackage{times}
\title{Transfer Learning for Customized Car Racing Environments}
\author{
Benedict Florance Arockiaraj [benarock@seas],
Richard Chang [rchang24@seas],
Wesley Yee [wesyee@seas]
}

\begin{document}

\begin{abstract}
Transfer Learning, a technique where a model/agent can use the knowledge/expertise that it gained from one task and exploit that to solve another closely-related task, is often used in tackling problems in deep learning. Through this project, we explore transfer learning in the purview of deep reinforcement learning. Specifically, we want to use transfer learning to achieve the fast lap times in OpenAI's Car racing environment by training the agent on one circuit, and racing it on other customized target environments by zero-shot transfer or by additional fine-tuning. In addition, we compare the performance of model-based and model-free approaches, and observe that model-based approaches dominate in performance and converge faster than model-free approaches in this environment. We observe that transfer learning in most setups not only boosts the performance on the target domain, but also shows high performance ability during learning.
\end{abstract}

\maketitle



\section{Introduction}
Reinforcement Learning (RL) is a go-to technique to solve tasks that involve decision making in a sequential manner and the agent learns to solve the task exploiting its interactions with the environment. The influx of deep-learning techniques have led to the use of deep learning models to build more expressive function approximators.  However, the most classic dilemma in RL is the tradeoff between exploration and exploitation. RL methods are data hungry, and the agent needs a large amount of explorative environment interactions to improve its performance. However, environments often tend to have sparse rewards, partial observability and high-dimensional continuous control spaces and thus acquiring a large number of environment interactions is costly. 

Transfer Learning comes to the rescue here and is a technique for transferring external expertise from a task/source domain and use it to advantage for the learning process of another task/target domain. It is widely used in the context of supervised deep learning especially for tasks like classification, object detection, segmentation etc. However, transfer learning in the context of deep-reinforcement learning is a bit non-trivial as the knowledge needs to transfer in the setting of an MDP and the expert knowledge can be of different forms and can be transferred in different ways. In this work, we explore transfer learning for OpenAI Gym's car racing environment \cite{brockman2016openai}, a continuous control task with pixel observations for top-down car-racing.

One classification of RL algorithms is how they make predictions about the next state and reward. In particular, model-free reinforcement learning obtains the optimal policy without estimating the dynamics (transition/reward function), while model-based methods `model' the environment and use this model to plan accordingly. In particular, we are interested in how model-based and model-free approaches compare in performing on transfer learning experiments and how they transfer to customized versions (changes in race map, acceleration, braking or friction) of the same environment. 

\subsection{Contributions}
Our project presents an extensive analysis of various transfer-learning experiments for model-based and model-free RL approaches. Specifically, we perform ablation experiments on the following setup:
\begin{enumerate}
    \item Testing model-free (PPO \cite{schulman2017proximal}, SAC \cite{haarnoja2018soft} and DDPG \cite{lillicrap2015continuous}) and model-based (Dreamer \cite{hafner2019dream}) on the standard OpenAI CarRacing environment
    \item Transferring pre-trained models for model-free (PPO \cite{schulman2017proximal} and SAC \cite{haarnoja2018soft}) and model-based (Dreamer \cite{hafner2019dream}) to customized environment with changes in vehicular dynamics or race tracks as follows:
    \begin{enumerate}
        \item different race map
        \item different acceleration parameters
        \item different braking parameters
        \item different friction parameters
    \end{enumerate}
\end{enumerate}
As expected, we notice that model-based approaches dominate performance, converge faster and are highly sample efficient as opposed to the model-free approaches in this environment. We observe that transfer learning in most setups not only boosts the performance on the target domain, but also shows high performance ability during learning indicating potential for safe reinforcement learning.

\section{Background}
\subsection{Transfer Learning}
As we saw earlier, transfer learning in the context of deep-reinforcement learning is a bit non-trivial as the knowledge needs to transfer in the setting of an MDP and the expert knowledge can be of different forms and can be transferred in different ways:
\begin{enumerate}
    \item \textbf{type of knowledge:} The transferred knowledge could take the form of a policy giving an action probability distribution, value, or action-value functions, or it could be a set of expert experiences. More importantly, transferring knowledge from one RL algorithm, say expert demonstrations, cannot work on another RL algorithm, say DDPG \cite{lillicrap2015continuous}.
    \item \textbf{difference between source and target domain:} The domains could differ in terms of the task goal, action space, observation space or parameters of the environment with the same set goal (like our case).
    \item \textbf{environment-accessibility in the target domain:} Querying the target domain could be expensive and the transfer learning is limited by how often we are allowed to interact with the target environment. While sometimes directly transferring to a new domain without target domain samples could show high-performance (\textit{zero-shot transfer}), a common technique used is to do \textit{few-shot transfer} to obtain sufficient number of environment interactions to converge faster on a target domain.
\end{enumerate}
\subsection{Reinforcement Learning Algorithms}
In this subsection, we briefly present the common classification of reinforcement-learning algorithms and explain the algorithms used in this work in short.
\subsubsection{Classification:}
\begin{enumerate}
    \item \textbf{model-free vs model-based:} A model-free algorithm determines the optimal policy without estimating the dynamics of the environment (transition function/rewards). Instead, it estimates a `value function' or a `policy' directly from interactions with the environment like Q-learning or policy gradients. On the contrary, model-based algorithms like policy and value iteration `model' the world (transition function/rewards) and use this model to estimate the policy. 
    \item \textbf{on-policy vs off-policy:} On-policy methods estimate the value of a policy (\textit{target/estimation policy}) and use the same policy for control/taking actions (\textit{behaviour policy}). On the other hand, off-policy have a behavior policy different from the target policy which gives the flexibility to use deterministic estimation policies while using explorative control policies.
\end{enumerate}
\subsubsection{Algorithms used in this work:}
We compare three model-free approaches (DDPG \cite{lillicrap2015continuous}, PPO \cite{schulman2017proximal} and SAC \cite{haarnoja2018soft}) to a model-based approach Dreamer \cite{hafner2019dream}. We choose algorithms that support continuous action space. 
\begin{enumerate}
    \item \textbf{DDPG \cite{lillicrap2015continuous}:} DDPG is a model-free, off-policy and actor-critic approach that combines stabilization tricks used for DQN \cite{mnih2013playing} (its discrete counterpart) like replay buffer, target network and gradient clipping, with deterministic policy gradient to design an algorithm for continuous actions.
    \item \textbf{PPO \cite{schulman2017proximal}:} PPO is a model-free, on-policy approach that incorporates ideas like multiple workers from A2C \cite{mnih2016asynchronous} and improvement of the actor through trust regions from TRPO \cite{schulman2015trust}, with an objective that the new policy should not be too far away from the old policy before update, and avoids large updates by clipping.
    \item \textbf{SAC \cite{haarnoja2018soft}:} SAC is a model-free, off-policy and actor-critic approach that integrates double Q-learning trick from TD3 \cite{fujimoto2018addressing}, an extension of DDPG into the objective of maximizing a trade-off between expected return and entropy to encourage exploration.
    \item \textbf{Dreamer \cite{hafner2019dream}:} Dreamer is a model-based approach that learns in `imagination' by learning the latent dynamics through a world-model, learns an actor-critic model by predicting latent trajectories and grows the dataset by executing the learned action model. It has the ability to learn long-horizon behaviors as the training purely happens in imagination by predicting both state and action values.  
\end{enumerate}

\section{Related Work}
Zhu et al. \cite{zhu2020transfer} presents a detailed survey and framework for analyzing transfer learning for deep-reinforcement learning. However, experimentation on how model-free methods (\cite{schulman2017proximal}, \cite{haarnoja2018soft}, \cite{schulman2015trust}, \cite{fujimoto2018addressing}, \cite{lillicrap2015continuous}, \cite{fujimoto2018addressing}) fare against model-based methods (\cite{hafner2019dream}) for transfer learning is limited and not much explored. Asawa et al. \cite{asawa2017using} present experimentation on transfer-learning between OpenAI games like Snake and PuckWorld. Given the dearth of experimentation in this area, we plan to perform a study on transfer-learning experiments between model-based and model-free approaches for the OpenAI Gym Car Racing environment \cite{brockman2016openai}.


\section{Approach}

\subsection{Car Racing Environment:}
We had initially wanted to utilize the F1Tenth Gym which was designed by the UPenn xLab \footnote{\href{https://github.com/f1tenth/f1tenth_gym/issues/43}{Issue link}} as it offered a lightweight framework with which to run our experiments in. However, after an initial investigation, we discovered that the F1Tenth Gym was not well maintained and also did not have the normal interface (state, reward, done, info) which is utilized by most modern reinforcement learning libraries. Thus, we decided to use the OpenAI Gym environment. \footnote{ \href{https://gym.openai.com/envs/CarRacing-v0/}{OpenAI Gym Car Racing Environment description link}}

\subsubsection{Description:} The car racing game which is played in this environment is solved once the agent consistently gets 900+ points. Three actions comprise the action space: steering (-1 is full left, +1 is full right), gas, and braking. The observation space consists of a grid world of 96x96 pixels. The reward function is -0.1 every time frame elapsed and +1000/N for every track tile visited, where N is the total number of tiles visited in the track (e.g, if the agent finished in 732 frames, the reward would be 1000 - 0.1*732). The path around the circuit is considered complete when the agent has reached 95 percent of the tiles on the track. The car starts at rest in the center of the road.

\subsection{Customizations:}
We decided to customize the environment in order to test how the various RL algorithms would each perform as compared to the out-of-the-box environment, which we used as our baseline.  The customizations which we made are listed below.

\subsubsection{Race Map/Circuit:} 
\begin{figure}%
    \centering
    \subfloat[\centering Source Map]{{\includegraphics[width=5cm]{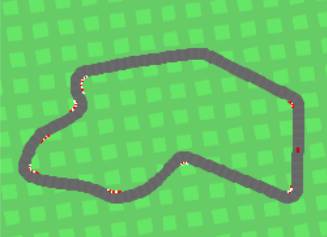} }}%
    \qquad
    \subfloat[\centering Target Map]{{\includegraphics[width=5cm]{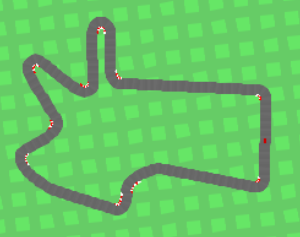} }}%
    \caption{Race Maps/Circuit}%
    \label{fig:map}%
\end{figure}

Since the CarRacing environment generates random maps for each game, this presented an issue as RL algorithms need to be trained on the same environment in order to learn. Thankfully, by specifying a random seed, we were able to generate two maps--a source map (used for training) and a target map (used for transfer learning evaluation). For our source environment, we chose a map which had less extreme turns, whereas the evaluation map contains much tighter turns as in figure \ref{fig:map}. This was intended to test whether the transfer learning algorithm would still perform well if presented with a more difficult target environment.

By default, the car racing environment uses keyboard controls to move the car along the circuit. While the agent can move around the track through keyboard control, this is by no means an optimal way to do so due to difficulties in precise control. Also, being able to drive on one circuit is no indication of being able to drive, and set a good lap time on another circuit. 

\subsubsection{Acceleration:} We set the max acceleration to the following values: +0.2/timestep. Default: +0.1/timestep.
\subsubsection{Braking:} We set the braking coefficient to the following values: 0.6. Default: 0.8. Lower braking coefficient indicates that the car needs a longer distance to come to a stop.
\subsubsection{Friction:} We set the grass friction coefficient to the following values: 0.8. Default: 0.6. Grass friction coefficient determines how much friction is between the wheels of the car and the grass. Higher grass friction coefficient indicates that the car skids less on grass and going through grass should not derail the lap as often.

\section{Experimental Results}
We ran each of the four algorithms on the source map, then for each of the algorithms, used zero-shot transfer learning to evaluate on the target map. For PPO, SAC, and Dreamer, we also tried few-shot transfer learning to learn the target map. In addition, transfer learning experiments were conducted from the base environment to new environments with modified acceleration, braking and friction parameters. We use a customized version of Dreamerv2 for training the Dreamer experiments. \footnote{\href{https://github.com/benedictflorance/dreamerv2_carracing}{Customized Dreamerv2 repo forked from Danijar Hafner's repository}} We use the Stable Baseline3 repository for our experiments in PPO, SAC and DDPG. \footnote{\href{https://github.com/DLR-RM/stable-baselines3}{Stable Baselines3 (SB3) Repository}} In addition, we wrote customized scripts to work on our different experimental scripts using the SB3 framework. \footnote{\href{https://github.com/benedictflorance/carracing-transfer-learning}{Customized scripts for SB3 (also has our work division report)}} Lastly, the base environment of Gym was modified and customized to our various source-target setups after extensive trial and error experimentation with the parameters. \footnote{\href{https://github.com/benedictflorance/gym}{Customized Gym repo for Car Racing environment forked from OpenAI's gym repository.}} Training on the base environment for all algorithms happened for 1M steps, while all transfer learning experiments were performed for 300k steps. We transfer the weights of the function approximators, i.e., the neural network weights in our transfer learning experiments. As expected, the training times were costly with 1M steps taking close to 24 hours of training.

\subsection{Performance on Standard Car Racing Environment (Base Environment):} 
\begin{figure}[ht]
    \centering
    \includegraphics[width=\textwidth]{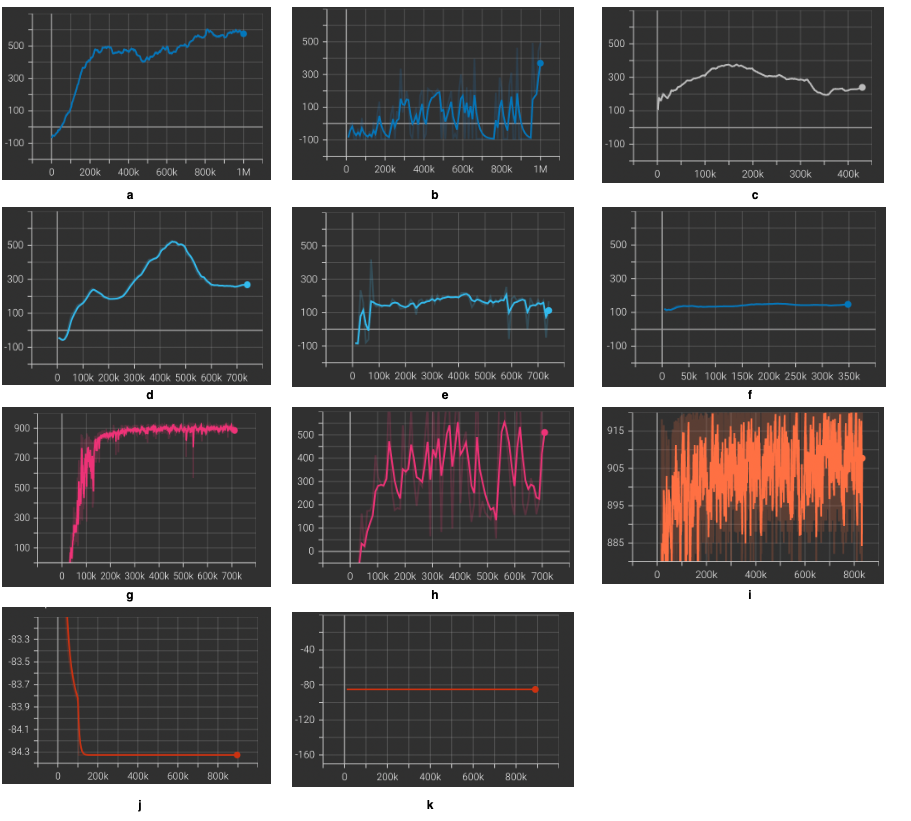}
    \caption{\textbf{Transfer Learning on different maps.} Maps show reward vs step count. a) PPO performance on source map b) PPO transfer from source to target map (zero-shot) c)PPO  transfer from source to target map (fine-tuned) d) SAC performance on source map e) SAC transfer from source to target map (zero-shot) f) SAC  transfer from source to target map (fine-tuned)  g) Dreamer performance on source map h) Dreamer transfer from source to target map (zero-shot) i) Dreamer  transfer from source to target map (fine-tuned) j) DDPG performance on source map k) DDPG transfer from source to target map (zero-shot) 
    }
    \label{fig:transfer_map}
\end{figure}

Figure \ref{fig:transfer_map} shows the plots for our experiments with source map environment, zero-shot transfer and fine-tuned performance on target environments. For PPO, SAC, and Dreamer, the agent was able to improve its performance on the source map, with PPO attaining a reward around 550 and SAC attaining a reward of 500 at 1M steps. The model-based algorithm, Dreamer, achieves the highest reward over 900 and converges drastically faster than the model-free counterparts. We see that Dreamer converges around 200k steps itself, far better than the model-free performance at that timestep. Meanwhile, DDPG struggled with this base environment, and after 900k steps, showed no meaningful change in reward. Moreover, the reward when running the DDPG algorithm was consistently around -84, and we struggled to make DDPG work in spite of various hyperparameter tuning (especially action noises like Normal noise/Ornstein Uhlenbeck noise).
In the context of hyperparameter tuning, we tuned the following hyperparameters for each of the tasks: (the best hyperparameter settings we found are mentioned in the scripts)
\begin{itemize}
    \item \textit{PPO}: Batch size, discount factor, learning rate, entropy coefficient, generalized advantage estimator tradeoff, value function coefficient, clipping parameter, max value for gradient clipping
    \item \textit{SAC}: Batch size, discount factor, learning rate, replay buffer size, Polyak update coefficient, training frequency, when to start learning
    \item \textit{DDPG}: Batch size, discount factor, learning rate, replay buffer size, gradient steps/rollout and action noise
    \item \textit{Dreamer:} Actor entropy, reward scaling, other parameters were set to DMC's defaults.
\end{itemize}

\subsection{Performance on different race circuit/map:} 
\subsubsection{Zero-shot Transfer:} For zero-shot transfer learning, PPO, SAC, and Dreamer were able to improve their performance across over several hundred thousand steps (PPO $\approx$ 400 and SAC $\approx$ 350), with Dreamer having the highest reward of around 500. However, for each of these three algorithms, the performance was quite below that experienced on the source map. As expected, DDPG still struggled just like it did on the source environment, with there being no meaningful changes in performance across 900k steps. So, we dropped DDPG from our further experimental setups.

\subsubsection{Finetuning:} While we fine-tuned for close to 300k steps, PPO and SAC showed similar performance as that of their zero-shot counterparts. Dreamer was able to perform the best, drastically better than the corresponding zero-shot transfer experiment, and performed at, or above the level, of the source map experiment (rewards $\approx$ 915) and thus solved the task on the target map. We didn't run DDPG for this experiment due to its poor performance on the base environment and the zero-shot transfer experiment.

\subsection{Performance on environment with different acceleration parameters:} 
\begin{figure}[ht]
    \centering
    \includegraphics[width=\textwidth]{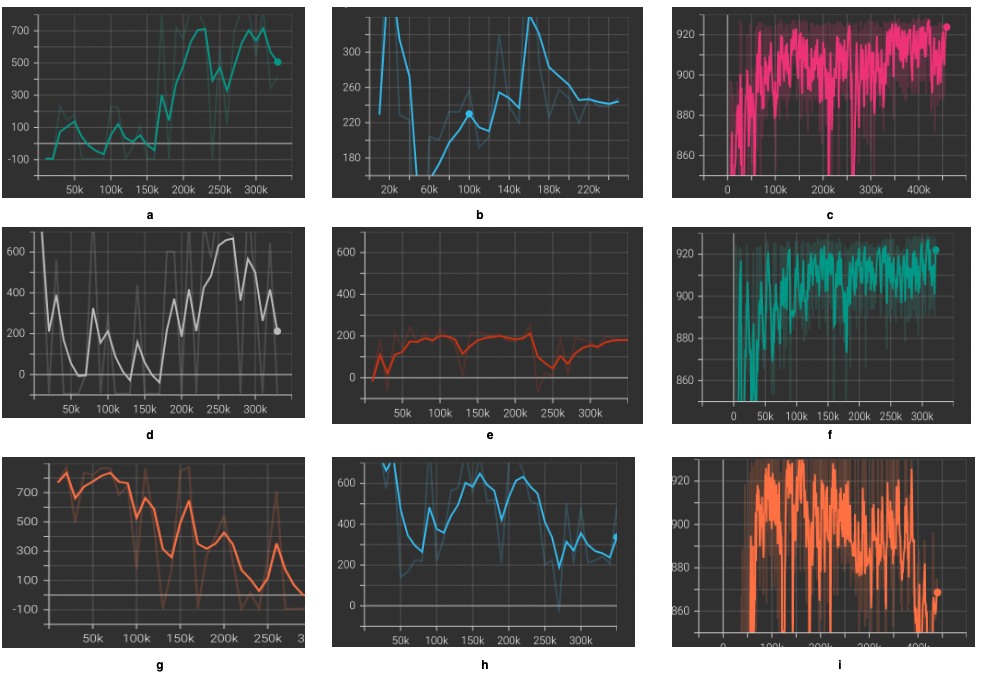}
    \caption{Maps show reward vs step count. Transfer Learning to a different \textbf{acceleration} setup: a) PPO performance b) SAC performance c) Dreamer performance. Transfer Learning to a different \textbf{braking} setup: d) PPO performance e) SAC performance f) Dreamer performance Transfer Learning to a different \textbf{friction } setup: g) PPO performance h) SAC performance i) Dreamer performance.
    }
    \label{fig:other}
\end{figure}

Figure \ref{fig:other} shows the performance of transfer learning experiments for PPO, SAC and Dreamer for customized Car Racing environments. When transferring onto an environment with a different acceleration, the performance of the agent when trained by PPO and Dreamer matched their performance on the base environment. The performance of SAC was slightly below that of the base environment.

\subsection{Performance on environment with different braking parameters:} Both the PPO and SAC algorithm performed worse on this new environment than on the base environment. The Dreamer algorithm performed at the same level as it did on the base environment, thus indicating the difficulty of the slow braking setup. 

\subsection{Performance on environment with different friction parameters:} The PPO algorithm performed significantly worse on this new environment than the base environment. Moreover, the PPO algorithm did worse as we finetuned for more timesteps, indicating that the PPO agent's goal to solve the task drifts in this new friction setup. The SAC algorithm performed at the same level as it did on the base environment. The Dreamer algorithm performed slightly worse than it did on the base environment, also indicating the difficulty of the high friction setup. However, we observe Dreamer trumps all other model-free algorithms in all the source-target setups.


\section{Discussion}
We recorded 20 second videos of the agent performing during the last timestep of each experiment setup for PPO, SAC and Dreamer agents. \footnote{\href{https://drive.google.com/drive/u/1/folders/1o9JxZ0eXth2tMaQjYiMq9A0hgWA0KXMw}{Link to gifs/videos of evaluation episodes in different setups/algos}} Looking at the videos, despite decent rewards for model-free setups, the algorithm may not have had the agent perform properly. For example, the agent may attempt to traverse the circuit in the incorrect direction such as in the SAC algorithm on the high acceleration environment. Overall, from the results and the videos, the model-free PPO algorithm outperformed the model-free SAC algorithm, and in some specific instances, performed better than, or at the level of the model-based Dreamer algorithm. We make some interesting observations comparing the gifs for Dreamer on the target map in zero-shot and finetuned setup. Since the target map had more turns than the source, the zero-shot transfer didn't help much and the car skidded to the grass on seeing a sharp U turn. After a while, it continues to move in the opposite direction. However, finetuning a bit fixes this, and the agent neatly learns to maneuver through the sharp U turn perfectly as seen with the Dreamer algorithm, indicating the power of transfer learning.

However, our major observations out of our work are four fold: 
\begin{enumerate}
    \item Model-free algorithms have higher sample complexity than model-based (can only attain close to 700 reward at 1M steps, while Dreamer solves the task in 200k iterations). 
    \item Thereby, model-free algorithms have brittle convergence properties and this requires meticulous hyperparameter tuning (we spent a lot of time tuning the hyperparameters for model-free algorithms, and still weren't able to get DDPG to work. Even with cherry-picked hyperparameters, there was no guarantee that the model-free algorithms would perform at the same level as it did on the base environment. However, Dreamer performed at the same level as it did on the base environment just with the default DMC hyperparameter settings).
    \item Model-based algorithm Dreamer showed the strongest performance on all experimental setups on customized environments (different maps, acceleration, braking or friction) in terms of data-efficiency, computation time, and final performance.
    \item In most of the setups, transfer learning, especially finetuning, helps to get the knack of solving the goal in the new target environment quickly. And in many setups, it matches, or outperforms the base environment.
\end{enumerate}

In conclusion, we were able to solve the task of car racing in customized environment setups, especially with model-based RL. Transfer learning results are exciting, as they not only improve performance and convergence, but this transfer has a lot of impact in performing safe reinforcement learning.  We were constrained by the amount of compute that we had as each experiment took a whole day to run. Given further time, we would be interested in experimenting on more complex environments. We would also love to extend our work to explore meta-RL and safe-RL in the context of car racing environments and/or test these algorithms in the real-world, outside of simulation.



{\small
\bibliographystyle{ieee_fullname}
\bibliography{bibfile}
}

\end{document}